\documentclass[11pt,a4paper]{article}
\usepackage[hyperref]{eacl2021}
\usepackage{times}
\usepackage{latexsym}

\usepackage{times}
\usepackage{url}
\usepackage{latexsym}
\usepackage{danudefs}
\usepackage{multirow}

\usepackage{microtype}

\aclfinalcopy 

\usepackage[english]{babel}
\usepackage{hyperref}
\addto\captionsenglish{%
}
\addto\extrasenglish{%
}

\usepackage{esvect}
\usepackage{graphicx}
\usepackage{comment}
\usepackage{caption}
\usepackage{subcaption}
\usepackage{booktabs}

\usepackage{color}

\usepackage{todonotes}

\makeatletter
\if@todonotes@disabled

\else

\fi
\makeatother

\usepackage{booktabs}
\usepackage{graphics}
\usepackage{autobreak}
\usepackage{adjustbox}
\usepackage{breqn}

\title{Debiasing Pre-trained Contextualised Embeddings}

\author{
    Masahiro Kaneko\\
    Tokyo Metropolitan University \\
  {\tt kaneko-masahiro@ed.tmu.ac.jp}
    \And
    Danushka Bollegala\Thanks{ Danushka Bollegala holds concurrent appointments as a Professor at University of Liverpool and as an Amazon Scholar. This paper describes work performed at the University of Liverpool and is not associated with Amazon.} \\
  University of Liverpool, Amazon\\
  {\tt danushka@liverpool.ac.uk}}

\date{}

\begin{document}
\maketitle
\begin{abstract}
In comparison to the numerous debiasing methods proposed for the static non-contextualised word embeddings, the discriminative biases in contextualised embeddings have received relatively little attention.
We propose a fine-tuning method that can be applied at token- or sentence-levels to debias pre-trained contextualised embeddings.
Our proposed method can be applied to any pre-trained contextualised embedding model, without requiring to retrain those models.
Using gender bias as an illustrative example, we then conduct a systematic study using several state-of-the-art (SoTA) contextualised representations on multiple benchmark datasets to evaluate the level of biases encoded in different contextualised embeddings before and after debiasing using the proposed method.
We find that applying token-level debiasing for all tokens and across all layers of a contextualised embedding model produces the best performance.
Interestingly, we observe that there is a trade-off between creating an accurate vs. unbiased contextualised embedding model, and different contextualised embedding models respond differently to this trade-off.

\end{abstract}

\section{Introduction}


Contextualised word embeddings have significantly improved performance in numerous natural language processing (NLP) applications~\cite{BERT,liu2019roberta,Clark2020ELECTRAPT} and have established as the de facto standard for input text representations.
Compared to static word embeddings~\cite{Glove,Milkov:2013} that represent a word by a single vector in all contexts it occurs, contextualised embeddings use dynamic context dependent vectors for representing a word in a specific context.
Unfortunately however, it has been shown that, similar to their non-contextual counterparts, contextualised text embeddings also encode various types of unfair biases~\cite{Zhao:2019a,bordia-bowman-2019-identifying,may-etal-2019-measuring,Tan:2019,bommasani-etal-2020-interpreting,kurita-etal-2019-measuring}.
This is a worrying situation because such biases can easily propagate to the downstream NLP applications that use contextualised text embeddings.

Different types of unfair and discriminative biases such as gender, racial and religious biases have been observed in 
static word embeddings~\cite{Tolga:NIPS:2016,Zhao:2018aa,Rudinger:2018aa,Zhao:2018ab,Elazar:EMNLP:2018,kaneko-bollegala-2019-gender}.
 As discussed later in \autoref{sec:related} different methods have been proposed for debiasing static word embeddings such as projection-based methods~\cite{kaneko-bollegala-2019-gender,Zhao:2018ab,Tolga:NIPS:2016,Ravfogel:2020} and adversarial methods~\cite{Xie:NIPS:2017,gonen-goldberg-2019-lipstick}.
In contrast, despite multiple studies reporting that contextualised embeddings to be unfairly biased, methods for debiasing contextualised embeddings are relatively under explored~\cite{dev2020on-measuring,Nadeem:2020,Nangia:2020}.
Compared to static word embeddings, debiasing contextualised embeddings is significantly more challenging due to several reasons as we discuss next.

First, compared to static word embedding models where the semantic representation of a word is limited to a single vector, contextualised embedding models have a significantly large number of parameters related in complex ways.
For example, BERT-large model~\cite{BERT} contains 24 layers, 16 attention heads and 340M parameters. 
Therefore, it is not obvious which parameters are responsible for the unfair biases related to a particular word.
Because of this reason, projection-based methods, popularly used for debiasing pre-trained static word embeddings, cannot be directly applied to debias pre-trained contextualised word embeddings.

Second, in the case of contextualised embeddings, the biases associated with a particular word's representation is a function of both the target word itself and the context in which it occurs.
Therefore, the same word can show unfair biases in some contexts and not in the others.
It is important to consider the words that co-occur with the target word in different contexts when debiasing a contextualised embedding model.

Third, pre-training large-scale contextualised embeddings from scratch is time consuming and require specialised hardware such as GPU/TPU clusters.
On the other hand, fine-tuning a pre-trained contextualised embedding model for a particular task (possibly using labelled data for the target task) is relatively less expensive.
Consequently, the standard practice in the NLP community has been to share\footnote{\url{https://huggingface.co/transformers/pretrained_models.html}} pre-trained contextualised embedding models and fine-tune as needed.
Therefore, it is desirable that a debiasing method proposed for contextualised embedding models can be applied as a fine-tuning method.
In this view, counterfactual data augmentation methods~\cite{Zmigrod:2019,hall-maudslay-etal-2019-name,Zhao:2019a} that swap gender pronouns in the training corpus for creating a gender balanced version of the training data are less attractive when debiasing contextualised embeddings because we must retrain those models on the balanced corpora, which is more expensive compared to fine-tuning.

Using gender-bias as a running example, we address the above-mentioned challenges by proposing a debiasing method that fine-tunes pre-trained contextualised word embeddings\footnote{Code and debiased embeddings: \url{https://github.com/kanekomasahiro/context-debias}}.
Our proposed method retains the semantic information learnt by the contextualised embedding model with respect to gender-related words, while simultaneously removing any stereotypical biases in the pre-trained model.
In particular, our proposed method is agnostic to the internal architecture of the contextualised embedding method and we apply  it to debias different pre-trained embeddings such as BERT, RoBERTa~\cite{liu2019roberta}, ALBERT~\cite{Lan2020ALBERTAL}, DistilBERT~\cite{Sanh2019DistilBERTAD} and ELECTRA~\cite{Clark2020ELECTRAPT}.
Moreover, our proposed method can be applied at token-level or at sentence-level, enabling us to debias at different granularities and on different layers in the pre-trained contextualised embedding model.

Following prior work, we compare the proposed debiasing method in two sentence-level tasks:  
Sentence Encoder Association Test~\cite[\textbf{SEAT};][]{may-etal-2019-measuring} and Multi-genre co-reference-based Natural Language Inference~\cite[\textbf{MNLI};][]{dev2020on-measuring}.
Experimental results show that the proposed method not only debiases all contextualised word embedding models compared,
 but also preserves useful semantic information for solving downstream tasks such as sentiment classification~\cite{socher-etal-2013-recursive}, paraphrase detection~\cite{dolan-brockett-2005-automatically}, semantic textual similarity measurement~\cite{cer-etal-2017-semeval}, natural language inference~\cite{10.1007/11736790_9,article11111} and solving Winograd schema~\cite{10.5555/3031843.3031909}.
 We consider gender bias as a running example throughout this paper and evaluate the proposed method with respect to its ability to overcome gender bias in contextualised word embeddings, and defer extensions to other types of biases to future work.

%


\section{Related Work}
\label{sec:related}

Prior work on debiasing word embeddings can be broadly categorised into two groups depending on whether they consider static or contextualised word embeddings. 
Although we focus on contextualised embeddings in this paper, we first briefly describe prior work on debiasing static embeddings for completeness of the discussion.
\paragraph{Bias in Static Word Embeddings:} 
\newcite{Tolga:NIPS:2016} proposed a post-processing approach that projects gender-neutral words into a subspace, which is orthogonal to the gender direction defined by a list of gender-definitional words.
However, their method ignores gender-definitional words during the subsequent debiasing process, and focus only on words that are \emph{not} predicted as gender-definitional by a classifier. 
Therefore, if the classifier erroneously predicts a stereotypical word as gender-definitional, it would not get debiased.
\newcite{Zhao:2018ab} modified the original GloVe~\cite{Glove} objective to learn gender-neutral word embeddings (GN-GloVe) from a given corpus.
Unlike the above-mentioned methods, \newcite{kaneko-bollegala-2019-gender} proposed GP-GloVe, a post-processing method to preserve gender-related information with autoencoder~\cite{kaneko-bollegala-2020-autoencoding}, while removing discriminatory biases from stereotypical cases.

Adversarial learning~\cite{Xie:NIPS:2017,Elazar:EMNLP:2018,Li:2018ab} for debiasing first encode the inputs and then two classifiers are jointly trained -- one predicting the target task (for which we must ensure high prediction accuracy) and the other for protected attributes (that must not be easily predictable). 
\newcite{Elazar:EMNLP:2018} showed that although it is possible to obtain chance-level development-set accuracy for the protected attributes during training, a post-hoc classifier trained on the encoded inputs can still manage to reach substantially high accuracies for the protected attributes. 
They conclude that adversarial learning alone does not guarantee invariant representations for the protected attributes.
\newcite{Ravfogel:2020} found that iteratively projecting word embeddings to the null space of the gender direction to further improve the debiasing performance. 
\paragraph{Benchmarks for biases in Static Embeddings:}
Word Embedding Association Test~\cite[WEAT;][]{Caliskan2017SemanticsDA} quantifies various biases (e.g. gender, race and age) using semantic similarities between word embeddings.
Word Association Test (WAT) measures gender bias over a large set of words~\cite{du-etal-2019-exploring} by calculating the gender information vector for each word in a word association graph created in the Small World of Words project~\cite[SWOWEN;][]{Deyne2019TheW} by propagating masculine and feminine words via a random walk~\cite{Zhou2003LearningWL}.
SemBias dataset~\cite{Zhao:2018ab} contains three types of word-pairs: 
(a) \textbf{Definition}, a gender-definition word pair (e.g. hero -- heroine), 
(b) \textbf{Stereotype}, a gender-stereotype word pair (e.g., manager -- secretary) 
and (c) \textbf{None}, two other word-pairs with similar meanings unrelated to gender (e.g., jazz -- blues, pencil -- pen). 
It uses the cosine similarity between the gender directional vector, $(\vv{he} - \vv{she})$, and the offset vector $(\vec{a} - \vec{b})$ for each word pair, $(a,b)$, in each set to measure gender bias.
WinoBias~\cite{Zhao:2018aa} uses the ability to predict gender pronouns with equal probabilities for gender neutral nouns such as occupations as a test for the gender bias in embeddings. 
\paragraph{Bias in Contextualised Word Embeddings:}
\newcite{may-etal-2019-measuring} extended WEAT using templates to create a sentence-level benchmark for evaluating bias called SEAT. In addition to the attributes proposed in WEAT, they proposed two additional bias types: \emph{angry black woman} and \emph{double binds} (when a woman is doing a role that is typically done by a man that woman is seen as arrogant). They show that compared to static embeddings, contextualised embeddings such as BERT, GPT and ELMo are less biased.
However, similar to WEAT, SEAT also only has positive predictive ability and cannot detect the absence of a bias.
\newcite{bommasani-etal-2020-interpreting} evaluated the bias in contextualised embeddings by first distilling static embeddings from contextualised embeddings and then using WEAT tests for different types of biases such as gender (male, female),  racial (White, Hispanic, Asian) and religion (Christianity, Islam). They found that aggregating the contextualised embedding of a particular word in different contexts via averaging to be the best method for creating a static embedding from a contextualised embedding.

\newcite{Zhao:2019a} showed that contextualised ELMo embeddings also learn gender biases present in the training corpus. 
Moreover, these biases propagate to a downstream coreference resolution task. 
They showed that data augmentation by swapping gender helps more than neutralisation by a projection. 
They obtain the embedding of two input sentences with reversed gender from ELMo, and obtain the debiased embedding by averaging them.
It can only be applied to feature-based embeddings, so it cannot be applied to fine-tuning based embeddings like BERT.
We directly debias the contextual embeddings.
Additionally, data augmentation requires re-training of the embeddings, which is often costly compared to fine-tuning.
\newcite{kurita-etal-2019-measuring} created masked templates such as ``\_\_ is a nurse'' and used BERT to predict the masked gender pronouns. 
They used the log-odds between male and female pronoun predictions as an evaluation measure and showed that BERT to be biased according to it.
\newcite{karve-etal-2019-conceptor} learnt conceptor matrices using class definitions in the WEAT and used the negated conceptors to  debias ELMo and BERT.
Although their method was effective for ELMo, the results on BERT were mixed.
This method can only be applied to context-independent vectors, and it requires the creation of static embeddings from BERT and ELMo as a pre-processing step for debiasing the context-dependent vectors.
Therefore, we do not compare against this method in the present study, where we evaluate on context-dependent vectors. 

\newcite{dev2020on-measuring} used natural language inference (NLI) as a bias evaluation task, where the goal is to ascertain if one sentence (i.e. premise) entails or contradictions another (i.e. hypothesis), or if neither conclusions hold (i.e. neutral).
The premise-hypothesis pairs are constructed to elicit various types of discriminative biases.
They showed that orthogonal projection to gender direction~\cite{DBLP:conf/aistats/DevP19} can be used to debias contextualised embeddings as well.
However, their method can be applied only to the noncontextualised layers (ELMo's Layer 1 and BERT's subtoken layer).
In contrast, our proposed method can be applied to \emph{all} layers in a contextualised embedding and outperforms their method on the same NLI task.
And our debiasing approach does not require task-dependent data.


\section{Debiasing Contextualised Embeddings}
\label{sec:method}

\begin{figure}
  \centering
  \includegraphics[width=7cm]{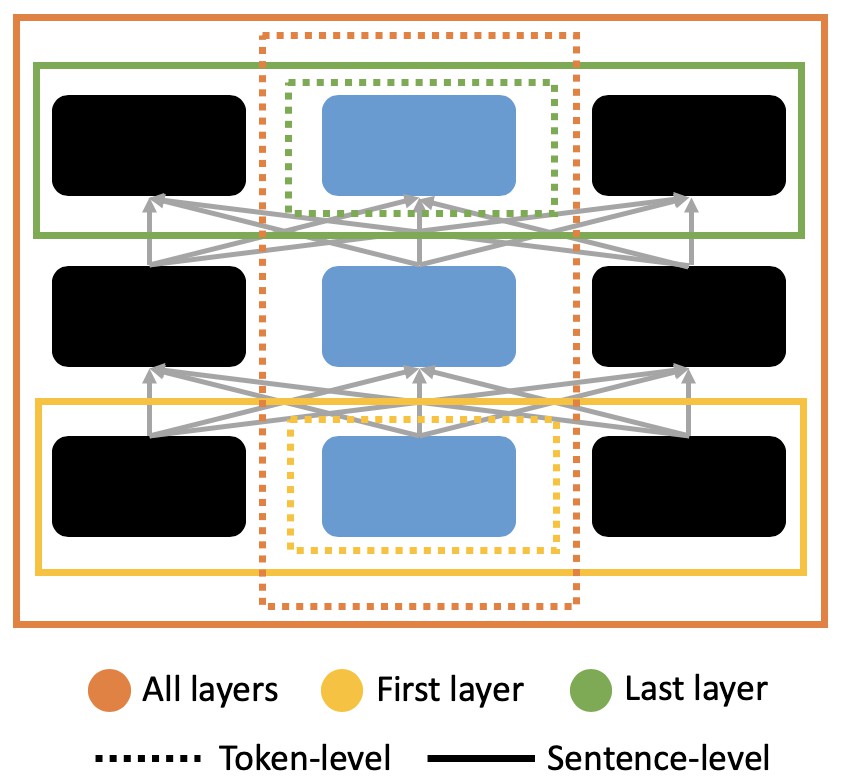}
  \caption{Types of hidden states in $E$ considered in the proposed method. The blue boxes in the middle correspond to the hidden states of the target token.}
  \label{fig:method}
\end{figure}

We propose a method for debiasing pre-trained contextualised word embeddings in a fine-tuning setting that simultaneously
(a) preserves the semantic information in the pre-trained contextualised word embedding model, and
(b) removes discriminative gender-related biases via an orthogonal projection in the intermediate (hidden) layers
by operating at token- or sentence-levels.
Fine-tuning allows debiasing to be carried out without requiring large amounts of tarining data or computational resources.
Our debiasing method is independent of model architectures or their pre-training methods, and can be adapted to a wide range of contextualised embeddings 
as shown in \autoref{sec:exp:models}.

Let us define two types of words: \emph{attribute} words ($\cV_a$)  and \emph{target} words ($\cV_t$).
For example, in the case of gender bias, attribute words consist of multiple word sets such as feminine (e.g. she, woman, her) and masculine (e.g. he, man, him) words, whereas target words can be occupations (e.g. doctor, nurse, professor), which we expect to be gender neutral.
We then extract sentences that contain an attribute or a target word.
Sentences contain more than one attribute (or target) words are excluded to avoid ambiguities.
Let us denote the set of sentences extracted for an attribute or a target word $w$ by $\Omega(w)$.
Moreover, let $\cA = \bigcup_{w \in \cV_a} \Omega(w)$ and $\cT = \bigcup_{w \in \cV_t} \Omega(w)$ be the sets of sentences containing respectively all of the attribute and target words.
We require that the debiased contextualised word embeddings preserve semantic information w.r.t. the sentences in $\cA$, and remove any discriminative biases w.r.t. the sentences in $\cT$.


Let us consider a contextualised word embedding model $E$, with pre-trained model parameters $\vec{\theta}_e$.
For an input sentence $x$, let us denote the embedding of token $w$ in the $i$-th layer of $E$ by $E_{i}(w, x; \vec{\theta}_e)$.
Moreover, let the total number of layers in $E$ to be $N$.
In our experiments, we consider different types of encoder models such as $E$.
To formalise the requirement that the debiased word embedding $E_{i}(t, x; \vec{\theta}_e)$ of a target word $t \in \cV_t$ must not contain any information related to a protected attribute $a$, we consider the inner-product between the noncontextualised embedding $\vec{v}_{i}(a)$ of $a$ and $E_{i}(t, x; \vec{\theta}_e)$ as a loss $L_{i}$ given by \eqref{eq:loss-t}. 
\begin{align}
    \label{eq:loss-t}
    L_{i} \!=\!\!\!\sum_{t \in \cV_t}\sum_{x \in \Omega(t)} \sum_{a \in \cV_a} \left( {\vec{v}_{i}(a)}\T E_{i}(t, x; \vec{\theta}_e) \right)^2
\end{align}
Here, $\vec{v}_{i}(a)$ is computed by averaging the contextualised embedding of $a$ in the $i$-th layer of $E$ over all sentences in $\Omega(a)$ following \newcite{bommasani-etal-2020-interpreting} and is given by \eqref{eq:v}.
\begin{align}
    \label{eq:v}
    \vec{v}_{i}(a) = \frac{1}{|\Omega(a)|} \sum_{x \in \Omega(a)} E_{i}(a, x; \vec{\theta}_e)
\end{align}
Here, $|\Omega(a)|$ denotes the total number of sentences in $\Omega(a)$.
If a word is split into multiple sub-tokens, we compute the contextualised embedding of the word by averaging the contextualised embeddings of its constituent sub-tokens.
Minimising the loss $L_{i}$ defined by \eqref{eq:loss-t} with respect to $\vec{\theta}_e$ forces the hidden states of $E$ to be orthogonal to the protected attributes such as gender.

Although removing discriminative biases in $E$ is our main objective, we must ensure that simultaneously we preserve as much useful information that is encoded in the pre-trained model for the downstream tasks.
We model this  as a regulariser where we measure the squared $\ell_2$ distance between the contextualised word embedding of a word $w$ in the $i$-th layer in the original model, parametrised by $\vec{\theta}_{\textrm{pre}}$, and the debiased model as in \eqref{eq:reg}.
\par\nobreak
{\small
\begin{align}
    \label{eq:reg}
    L_{\textrm{reg}} \!=\!\! \sum_{x \in \cA}\sum_{w \in x} \sum_{i = 1}^{N} \norm{E_i(w,x;\vec{\theta}_e) - E_i(w,x;\vec{\theta}_{\textrm{pre}})}^2
\end{align}
}%
The overall training objective is then given by \eqref{eq:loss-total} as the linearly weighted sum of the two losses defined by \eqref{eq:loss-t} and \eqref{eq:reg}.
\begin{align}
 \label{eq:loss-total}
 L = \alpha L_{i} + \beta L_{\textrm{reg}}
\end{align}
Here, coefficients $\alpha, \beta \in [0,1]$ satisfy $\alpha + \beta = 1$.

As shown in \autoref{fig:method}, a contextualised word embedding model typically contains multiple layers.
It is not obvious which hidden states of $E$ are best for calculating $L_{i}$ for the purpose of debiasing.
Therefore, we compute $L_{i}$ for different layers in a particular contextualised word embedding model in our experiments.
Specifically, we consider three settings: debiasing only the \textbf{first} layer, \textbf{last} layer or \textbf{all} layers.
Moreover, $L_{i}$ can be computed only for the target words in a sentence $x$ as in \eqref{eq:loss-t}, or can be summed up for \emph{all} words in $w \in x$ (i.e. $\sum_{t \in \cV_t} \sum_{x \in \Omega(t)} \sum_{w \in x} \left(\vec{v}_i(a)\T E_i(w,x; \vec{\theta}_e)\right)^2$). 
We refer to the former as \textbf{token-level} debiasing and latter \textbf{sentence-level} debiasing.
Collectively this gives us six different settings for the proposed debiasing method, which we evaluate experimentally in \autoref{sec:exp:models}.

\section{Experiments}


\subsection{Datasets}
\label{sec:data}


\begin{table*}[t]
\scriptsize
\begin{adjustbox}{width=\linewidth}
\begin{tabular}{ccc | cccc | cccccc}
\toprule
Model & \emph{Layer} & \emph{Unit} & SEAT-6 & SEAT-7 & SEAT-8 & \#$\dag$ & SST-2 & MRPC & STS-B & RTE & WNLI & Avg     \\
\midrule
\multirow{7}{*}{BERT} & \multirow{2}{*}{all} & token & $0.68^\dag$ & -0.09 & $0.60^\dag$ & 2 & 92.1 & 85.6 & 83.1 & 60.0 & 53.5 & 74.9  \\
& & sent & $1.13^\dag$ & 0.34 & 0.12 & \textbf{1} & 91.9 & 82.6 & 80.0 & 54.2 & 40.8 & 69.9 \\
& \multirow{2}{*}{last} & token & $1.02^\dag$ & -1.18 & $0.47^\dag$ & 2 & 92.2 & 86.9 & 82.3 & 58.1 & 56.3 & 75.2 \\
& & sent & $1.51^\dag$ & -0.60 & $1.52^\dag$ & 2 & 92.3 & 84.6 & 82.9 & 62.1 & 56.3 & 75.6 \\
& \multirow{2}{*}{first} & token & $0.88^\dag$ & 0.33 & $0.86^\dag$ & 2 & 92.4 & 87.1 & 82.6 & 62.1 & 50.7 & 75.0 \\
& & sent & $0.94^\dag$ & 0.32 & $0.97^\dag$ & 2 & 91.9 & 86.1 & 83.0 & 63.9 & 46.5 & 74.3 \\
& \multicolumn{2}{c|}{original} & $1.04^\dag$ & 0.18 & $0.81^\dag$ & 2 & 92.8 & 86.7 & 82.4 & 60.6 & 56.3 & 75.8 \\
& \multicolumn{2}{c|}{random} & $1.16^\dag$ & -0.08 & -0.29 & \textbf{1} & 92.2 & 87.4 & 81.9 & 63.2 & 54.9 & \textbf{75.9} \\
\midrule
\multirow{6}{*}{RoBERTa} & \multirow{2}{*}{all} & token & $0.51^\dag$ & 0.15  & 0.02 & \textbf{1} & 78.1 & 81.6 & 73.7 & 53.8 & 56.3 & 68.7 \\
& & sent & $1.27^\dag$ & $0.86^\dag$ & $1.14^\dag$ & 3 & 80.3 & 82.8 & 74.4 & 50.9 & 56.3 & 68.9 \\
& \multirow{2}{*}{last} & token & $1.17^\dag$ & -0.60 & $0.45^\dag$ & 2 & 79.9 & 83.7 & 74.1 & 52.3 & 56.3 & 69.3 \\
& & sent & $0.98^\dag$ & $0.75^\dag$ & $0.87^\dag$ & 3 & 69.5 & 81.5 & 72.9 & 52.7 & 56.3 & 66.6 \\
& \multirow{2}{*}{first} & token & $1.15^\dag$ & 0.26 & $0.54^\dag$ & 2 & 77.8 & 81.1 & 74.5 & 54.5 & 56.3 & 68.8 \\
& & sent & $1.21^\dag$ & 0.32 & $0.50^\dag$ & 2 & 79.0 & 82.5 & 74.5 & 51.6 & 56.3 & 68.8 \\
& \multicolumn{2}{c|}{original} & $1.21^\dag$ & $1.34^\dag$ & $1.01^\dag$ & 3 & 93.8 & 91.2 & 89.8 & 71.8 & 56.3 & \textbf{80.6} \\
& \multicolumn{2}{c|}{random} & $1.39^\dag$ & $0.40^\dag$ & $0.39^\dag$ & 3 & 73.4 & 82.5 & 73.9 & 53.4 & 49.3 & 66.5 \\
\midrule
\multirow{7}{*}{ALBERT} & \multirow{2}{*}{all} & token & 0.16 & 0.02 & 0.18 & \textbf{0} & 78.1 & 80.5 & 67.5 & 54.9 & 56.3 & 67,5 \\
& & sent & 0.18 & -0.05 & -0.77 & \textbf{0} & 77.3 & 81.7 & 69.9 & 46.9 & 56.3 & 66.4 \\
& \multirow{2}{*}{last} & token & $0.83^\dag$ & -1.15 & -0.76 & 1 & 77.8 & 81.2 & 68.9 & 47.3 & 56.3 & 66.3 \\
& & sent & $0.69^\dag$ & -0.06 & -0.10 & 1 & 78.3 & 80.1 & 71.3 & 55.2 & 56.3 & 68,2 \\
& \multirow{2}{*}{first} & token & 0.09 & 0.28 & $0.97^\dag$ & 1 & 77.9 & 81.6 & 70.0 & 52.0 & 56.3 & 67,6 \\
& & sent & 0.25 & $0.60^\dag$ & $1.18^\dag$ & 2 & 75.9 & 81.3 & 70.1 & 53.1 & 54.9 & 67,1 \\
& \multicolumn{2}{c|}{original} & 0.30 & $0.48^\dag$ & $1.12^\dag$ & 2 & 92.2 & 89.9 & 87.7 & 70.0 & 56.3 & \textbf{79.2} \\
& \multicolumn{2}{c|}{random} & $0.41^\dag$ & 0.34 & $1.08^\dag$ & 2 & 78.2 & 79.9 & 71.8 & 47.3 & 56.3 & 66.7 \\
\midrule
\multirow{7}{*}{DistilBERT} & \multirow{2}{*}{all} & token & $0.70^\dag$ & -0.83 & -0.66 & \textbf{1} & 90.4 & 87.8 & 80.8 & 56.0 & 42.3 & 71.5 \\
& & sent & $1.34^\dag$ & $1.01^\dag$ & $0.97^\dag$ & 3 & 91.4 & 83.3 & 78.8 & 57.4 & 53.5 & \textbf{72.9} \\
& \multirow{2}{*}{last} & token & $1.11^\dag$ & -0.03 & $1.38^\dag$ & 2 & 90.9 & 88.5 & 80.3 & 55.6 & 38.0 & 70.7 \\
& & sent & $1.57^\dag$ & -1.34 & 0.27 & \textbf{1} & 90.8 & 90.2 & 80.9 & 58.5 & 43.7 & 72.8 \\
& \multirow{2}{*}{first} & token & $1.19^\dag$ & $0.59^\dag$ & $0.52^\dag$ & 3 & 90.8 & 90.8 & 80.4 & 55.2 & 38.0 & 71.0 \\
& & sent & $1.19^\dag$ & $0.60^\dag$ & $0.55^\dag$ & 3 & 91.1 & 90.9 & 80.1 & 55.2 & 36.6 & 70.8 \\
& \multicolumn{2}{c|}{original} & $1.26^\dag$ & 0.31 & $0.74^\dag$ & 2 & 90.8 & 89.3 & 80.6 & 56.0 & 38.0 & 70.9 \\
& \multicolumn{2}{c|}{random} & $1.35^\dag$ & $0.66^\dag$ & -0.25 & 2 & 91.1 & 89.1 & 80.5 & 56.3 & 40.8 & 71.6 \\
\midrule
\multirow{6}{*}{ELECTRA} & \multirow{2}{*}{all} & token & 0.33 & 0.10 & 0.15 & \textbf{0} & 90.3 & 87.7 & 79.4 & 52.7 & 57.7 & \textbf{73.6} \\
& & sent & $0.42^\dag$ & 0.21 & 0.33 & 1 & 90.7 & 87.1 & 79.5 & 52.3 & 54.9 & 72.9 \\
& \multirow{2}{*}{last} & token & $0.55^\dag$ & 0.07 & 0.24 & 1 & 90.8 & 87.3 & 79.8 & 51.6 & 46.5 & 71.2 \\
& & sent & $0.50^\dag$ & $0.42^\dag$ & $0.32^\dag$ & 3 & 90.5 & 87.3 & 80.1 & 54.5 & 40.8 & 70.6 \\
& \multirow{2}{*}{first} & token & 0.31 &  0.10 & 0.33 & \textbf{0} & 90.4 & 86.9 & 79.7 & 53.1 & 56.3 & 73.4 \\
& & sent & 0.29 & 0.22 & 0.30 & \textbf{0} & 90.4 & 87.6 & 79.7 & 53.4 & 56.3 & 73.5 \\
& \multicolumn{2}{c|}{original} & 0.16 & $0.46^\dag$ & 0.04 & 1 & 90.5 & 87.9 & 80.4 & 54.5 & 46.5 & 72.0 \\
& \multicolumn{2}{c|}{random} & $0.43^\dag$ & $0.49^\dag$ & -0.22 & 2 & 90.4 & 87.7 & 78.5 & 51.3 & 54.9 & 72.6 \\
\bottomrule
\end{tabular}
\end{adjustbox}
\caption{Gender bias of contextualised embeddings on SEAT. $\dag$ denotes significant bias effects at $\al < 0.01$. }
\label{tbl:seat_gleu}
\vspace{-4mm}
\end{table*}

We used SEAT~\cite{may-etal-2019-measuring} 6, 7 and 8 to evaluate gender bias.
We use NLI as a downstream evaluation task and use the Multi-Genre Natural Language Inference data~\cite[MNLI;][]{williams-etal-2018-broad} for training and development following \newcite{dev2020on-measuring}.
In NLI, the task is to classify a given hypothesis and premise sentence-pair as entailing, contradicting, or neutral.
We programmatically generated the evaluation set following \newcite{dev2020on-measuring} by filling occupation words and gender words in template sentences.
The templates take the form ``The \textsf{subject verb} a/an \textsf{object}.'' and the created sentence-pairs are assumed to be neutral.

We used the word lists created by \newcite{Zhao:2018ab} for the attribute list of feminine and masculine words.
As for the stereotype word list for target words, we use the list created by  \newcite{kaneko-bollegala-2019-gender}.
Using News-commentary-v15 corpus\footnote{\url{http://www.statmt.org/wmt20/translation-task.html}} was extract 
11023, 42489 and 34148 sentences respectively for Feminine, Masculine and Stereotype words. 
We excluded sentences with more than 128 tokens in training data.
We randomly sampled 1,000 sentences from each type of extracted sentences as development data.

We used the GLEU benchmark~\cite{wang-etal-2018-glue} to evaluate whether the useful information in the pre-trained embeddings is retrained after debiasing.
To evaluate the debiased models with minimal effects due to task-specific fine-tuning, we used the following small-scale training data: Stanford Sentiment Treebank~\cite[SST-2;][]{socher-etal-2013-recursive}, Microsoft Research Paraphrase Corpus~\cite[MRPC;][]{dolan-brockett-2005-automatically}, Semantic Textual Similarity Benchmark~\cite[STS-B;][]{cer-etal-2017-semeval}, Recognising Textual Entailment~\cite[RTE;][]{10.1007/11736790_9,article11111,10.5555/1654536.1654538,Bentivogli09thefifth}, and Winograd Schema Challenge~\cite[WNLI;][]{10.5555/3031843.3031909}.
We evaluate the performance of the contextualised embeddings on the corresponding development data.

\subsection{Hyperparameters}

We used BERT~\cite[\textbf{bert-base-uncased};][]{BERT}, RoBERTa~\cite[\textbf{roberta-base};][]{liu2019roberta}, ALBERT~\cite[\textbf{albert-base-v2};][]{Lan2020ALBERTAL}, DistilBERT~\cite[\textbf{distilbert-base-uncased};][]{Sanh2019DistilBERTAD} and ELECTRA~\cite[\textbf{electra-small-discriminator};][]{Clark2020ELECTRAPT} in our experiments.\footnote{We used \url{https://github.com/huggingface/transformers}}
DistilBERT has 6 layers and the others 12.
We used the development data in SEAT-6 for hyperparameter tuning.
The hyperparameters of the models, except the learning rate and batch size, are set to their default values as in \texttt{run\_glue.py}.
Using greedy search, the learning rate was set to 5e-5 and the batch size to 32 during debiasing.
Optimal values for $\alpha = 0.2$ and $\beta = 0.8$ were found by a greedy search in $[0,1]$ with $0.1$ increments.
For the GLEU and MNLI experiments, we set the learning rate to 2e-5 and the batch size to 16.
Experiments were conducted on a GeForce GTX 1080 Ti GPU.

\subsection{Debiasing vs. Preserving Information}
\label{sec:exp:models}

\autoref{tbl:seat_gleu} shows the results on SEAT and GLEU where \textbf{original} denotes the pre-trained contextualised models prior to debiasing.
We see that original models other than ELECTRA contain significant levels of gender biases.
Overall, the \textbf{all-token} method that conducts token-level debiasing across all layers performs the best.
Prior work has shown that biases are learned at each layer \cite{bommasani-etal-2020-interpreting} and it is important to debias all layers.
Moreover, we see that debiasing at token-level is more efficient compared to at the sentence-level.
This is because in token-level debiasing, the loss is computed only on the target word and provides a more direct debiasing update for the target word
than in the sentence-level debiasing, which sums the losses over all tokens in a sentence.

To test the importance of carefully selecting the target words considering the types of biases that we want to remove from the embeddings,
we implement a \textbf{random} baseline where we randomly select target and attribute words from $\cV_{a} \cup \cV_{t}$ and perform \textbf{all-token} debiasing.
We see that \textbf{random} debiases BERT to some extent but is not effective on other models.
This result shows that the proposed debiasing method is \emph{not} merely a regularisation technique that imposes constraints on any arbitrary set of words,
but it is essential to carefully select the target words used for debiasing.

The results on GLEU show that BERT, DistilBERT and ELECTRA compared to the \textbf{original} embeddings, the debiased embeddings report comparable performances in most settings. 
This confirms that the proposed debiasing method preserves sufficient semantic information contained in the \textbf{original} embeddings that can be used to learn accurate prediction models for the downstream NLP tasks.\footnote{Although on WNLI \textbf{all-token} debiasing improves performance for DistilBERT and ELECTRA compared to the respective \textbf{original} models, this is insignificant as WNLI contains only 146 test instances.}
However, the performance of RoBERTa and ALBERT decrease significantly compared to their \textbf{original} versions after debiasing.
We suspect that these models are more sensitive to fine-tuning and hence lose their pre-trained information during the debiasing process.
We defer the development of techniques to address this issue to future research.

\subsection{Measuring Bias with Inference}

\begin{table}[t]
\begin{adjustbox}{width=\columnwidth,center}
\centering
\begin{tabular}{lccccc}
\toprule
Model & MNLI-m & MNLI-mm & NN & FN & T:0.7 \\
\midrule
\newcite{dev2020on-measuring} & \textbf{80.8} & 81.1 & 85.5 & \textbf{97.3} & 88.3 \\
all-token & 80.7 & \textbf{81.2} & \textbf{87.8} & 96.8 & \textbf{89.3} \\
original & \textbf{80.8} & 81.0 & 82.3 & 96.4 & 83.2 \\
random & 80.5 & 81.1 & 85.8 & 96.4 & 87.0  \\
\bottomrule
\end{tabular}
\end{adjustbox}
\caption{Debias results for BERT in MNLI.}
\label{tbl:snli_size}
\end{table}

\begin{table}[t]
\begin{adjustbox}{width=\columnwidth,center}
\centering
\begin{tabular}{lcccc}
\toprule
Model & \textit{Layer}          & SEAT-6 & SEAT-7 & SEAT-8 \\
\midrule
\multirow{3}{*}{BERT} & all     & \textbf{0.44} & 0.25 & \textbf{0.46} \\
                      & last    & 0.56 & \textbf{0.12} & 0.47 \\
                      & first   & 0.52 & 0.22 & 0.49 \\
\midrule
\multirow{3}{*}{RoBERTa}    & all       & \textbf{0.59} & \textbf{0.23} & 0.61  \\
                            & last      & 0.73 & 0.24 & 0.65  \\
                            & first     & 0.69 & 0.28 & \textbf{0.59}  \\
\midrule
\multirow{3}{*}{ALBERT}     & all       & \textbf{0.46} & 0.48 & \textbf{0.24}  \\
                            & last      & 1.15 & \textbf{0.26} & 0.60  \\
                            & first     & 0.54 & 0.89 & 0.95  \\
\midrule
\multirow{3}{*}{DistilBERT}     & all       & \textbf{0.66} & \textbf{-0.16} & 0.37 \\
                                & last      & 0.88 & 0.19  & \textbf{0.35} \\
                                & first     & 0.90 & 0.40  & 0.52 \\
\midrule
\multirow{3}{*}{ELECTRA}    & all       & \textbf{0.21} & \textbf{0.02} & \textbf{0.18}  \\
                            & last      & 0.34 & 0.20 & 0.21  \\
                            & first     & 0.28 & 0.13 & 0.34 \\
\bottomrule
\end{tabular}
\end{adjustbox}
\caption{Averaged scores over all layers in an embedding debiased at token-level,  measured on SEAT tests.}
\label{tbl:layer_seat}
\end{table}

\begin{figure*}[t!]
    \centering
	\begin{subfigure}[b]{0.3\textwidth}
		\centering
		\includegraphics[height=1.35in]{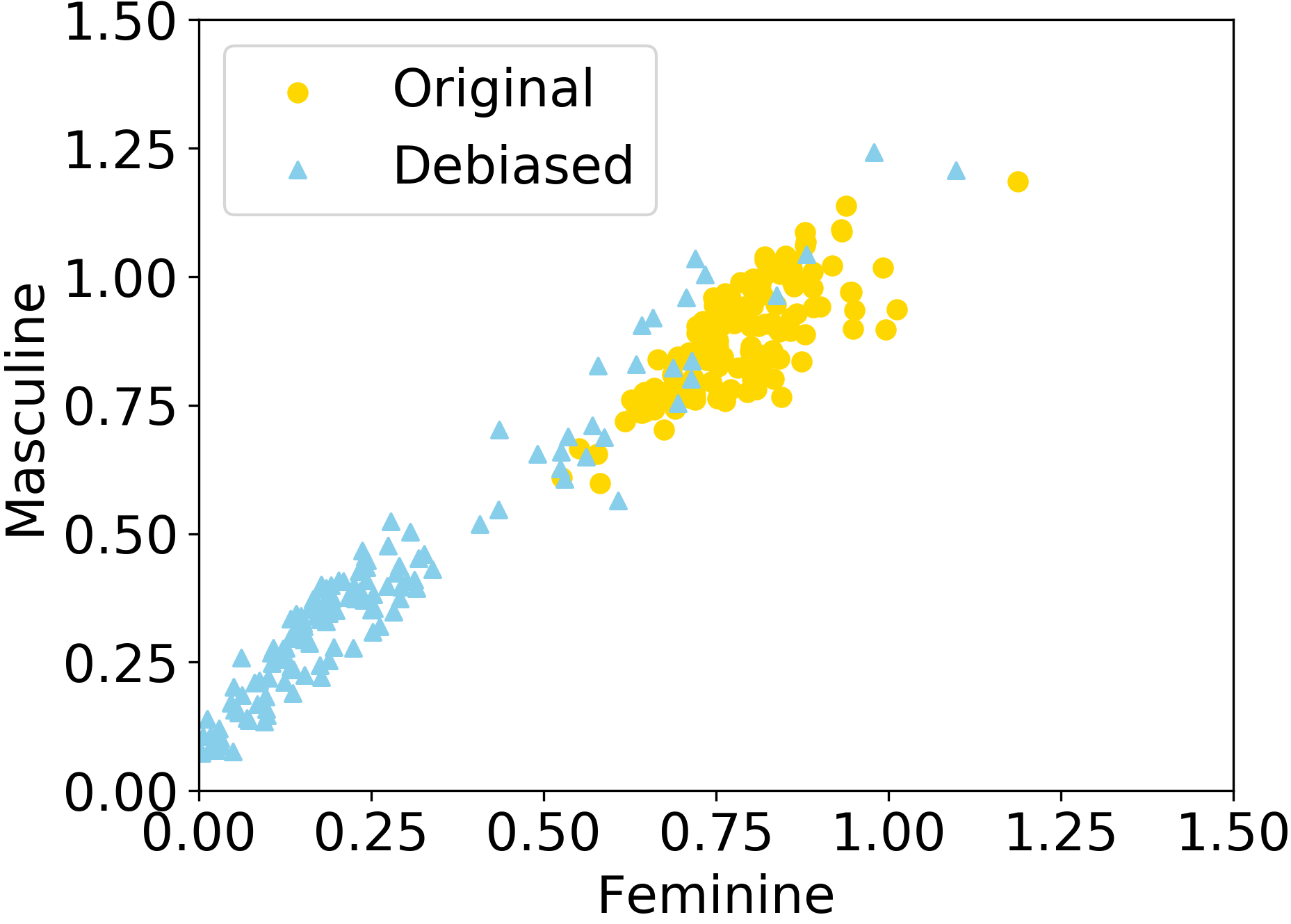}
		\caption{BERT}
		\label{fig:glove}
	\end{subfigure}
	\begin{subfigure}[b]{0.3\textwidth}
		\centering
		\includegraphics[height=1.35in]{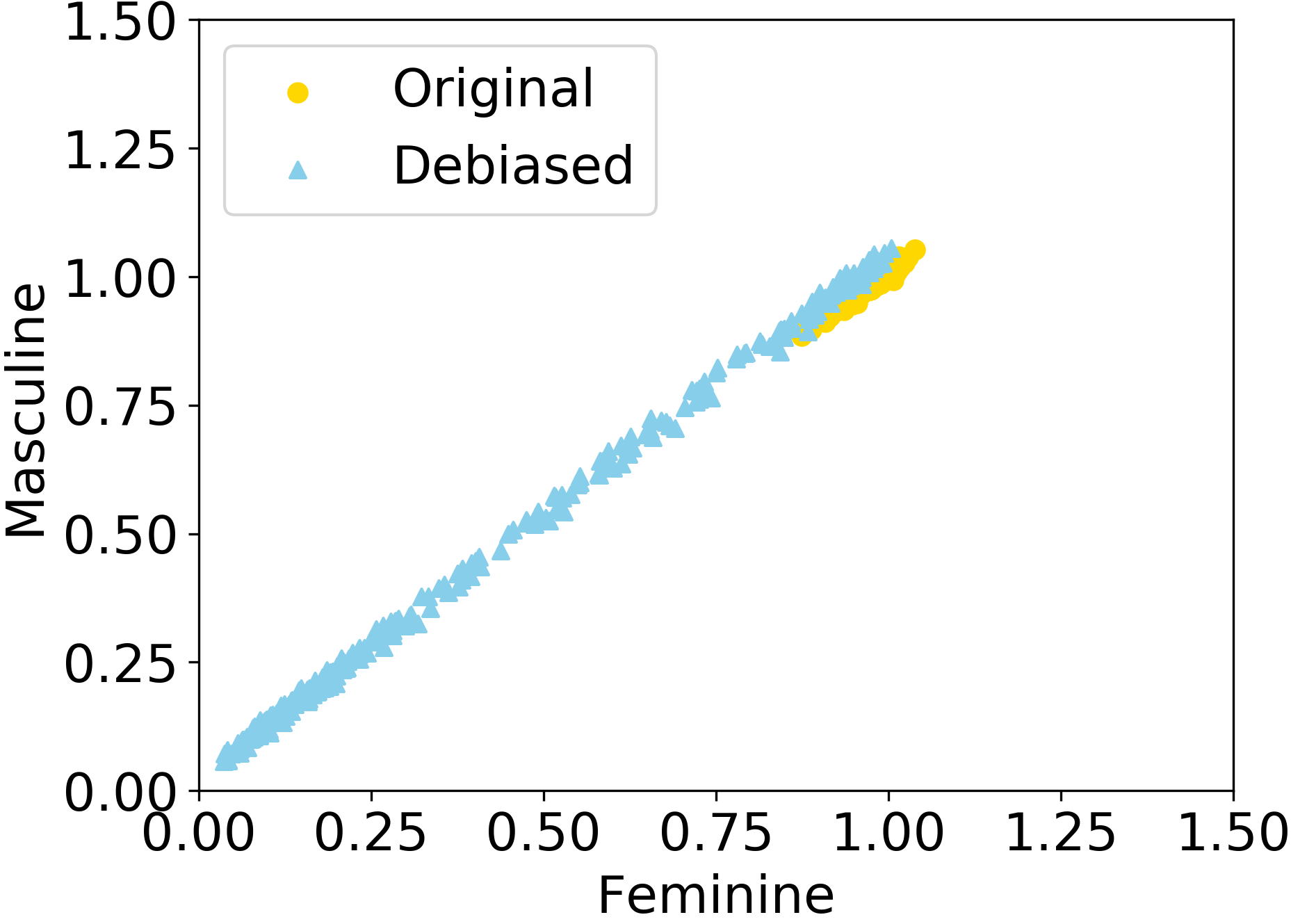}
		\caption{RoBERTa}
	\end{subfigure}
	\begin{subfigure}[b]{0.3\textwidth}
		\centering
		\includegraphics[height=1.35in]{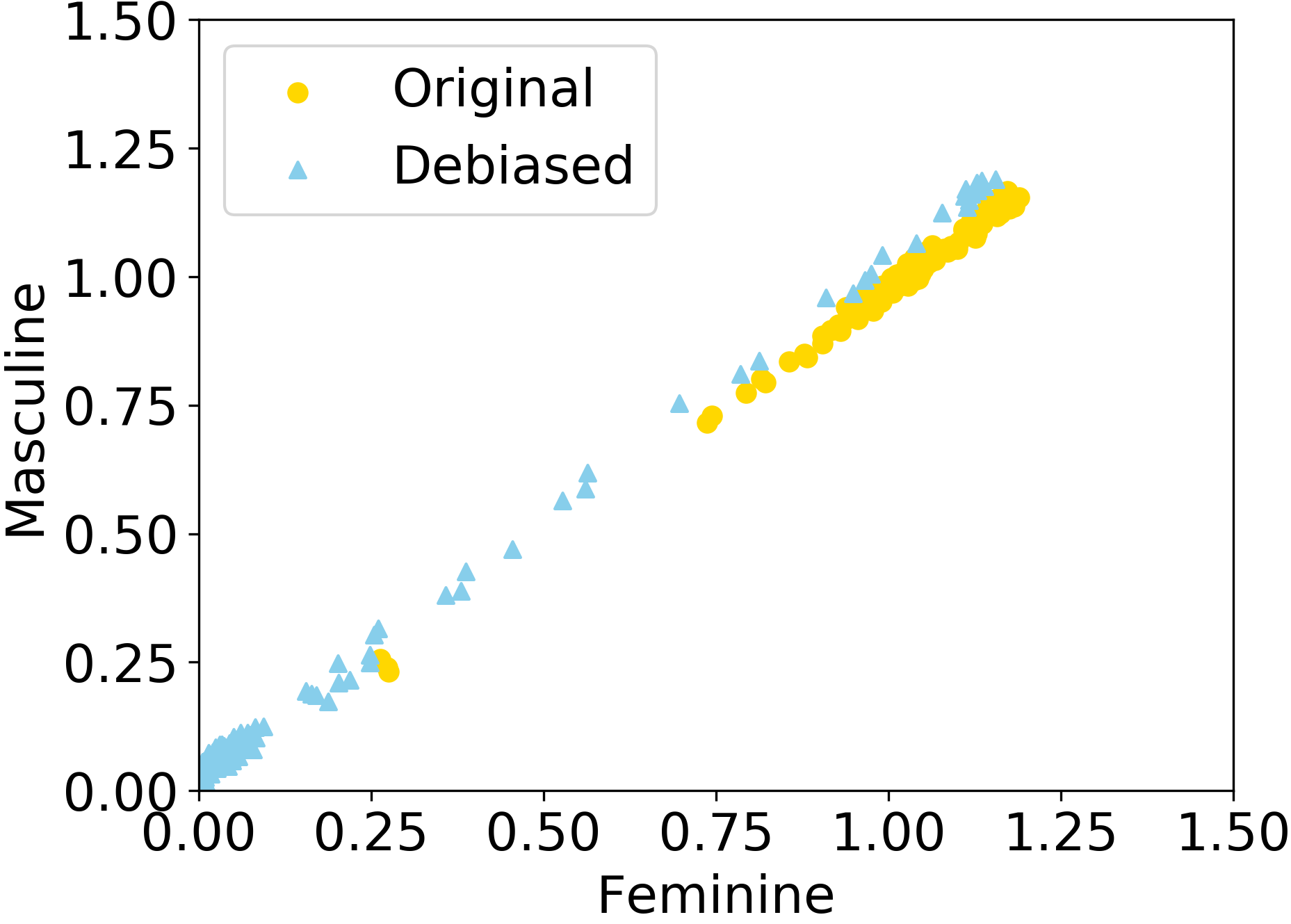}
		\caption{ALBERT}
	\end{subfigure}
	\vskip\baselineskip
	\begin{subfigure}[b]{0.3\textwidth}
		\centering
		\includegraphics[height=1.35in]{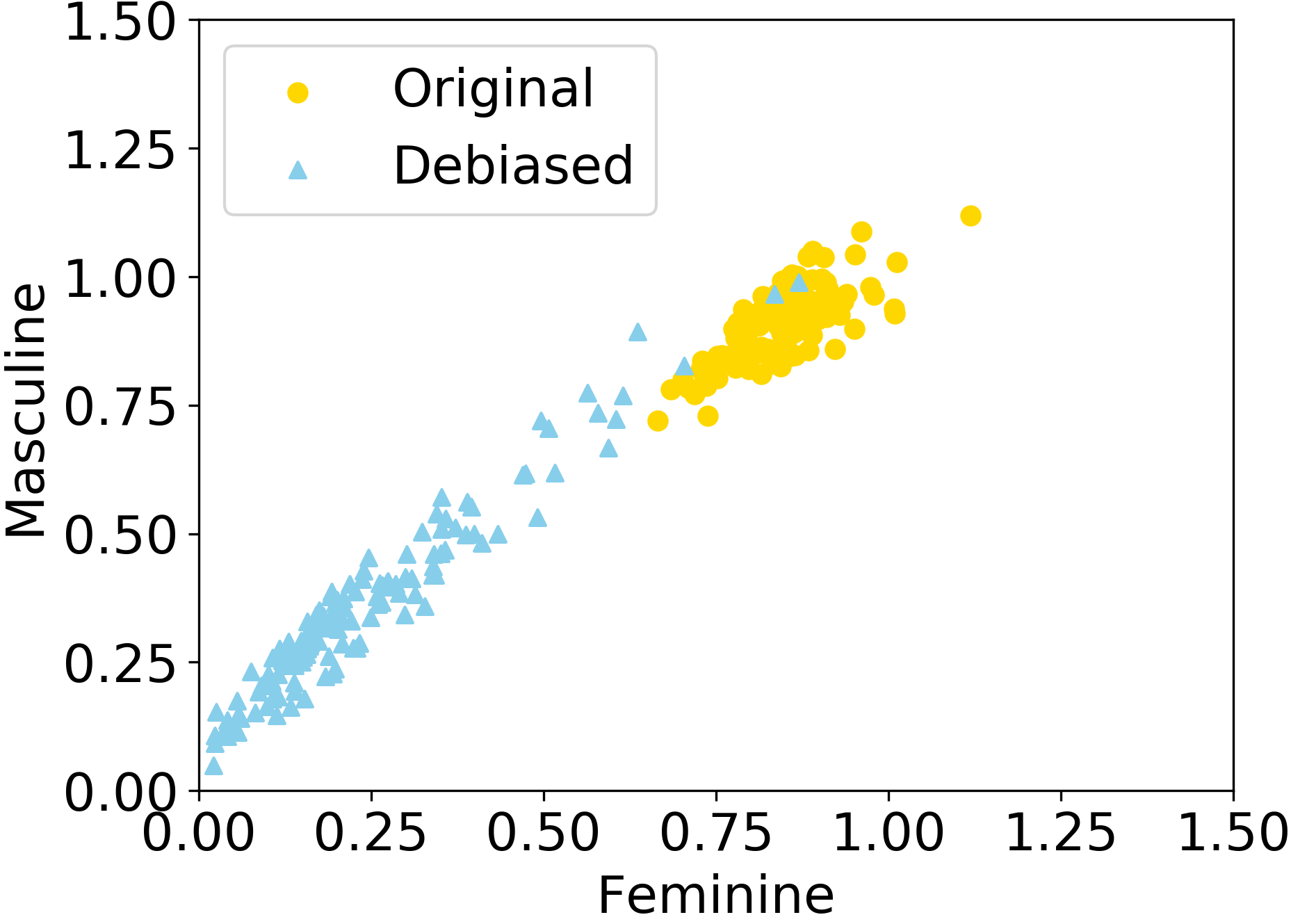}
		\caption{DistilBERT}
	\end{subfigure}
	\begin{subfigure}[b]{0.3\textwidth}
		\centering
		\includegraphics[height=1.35in]{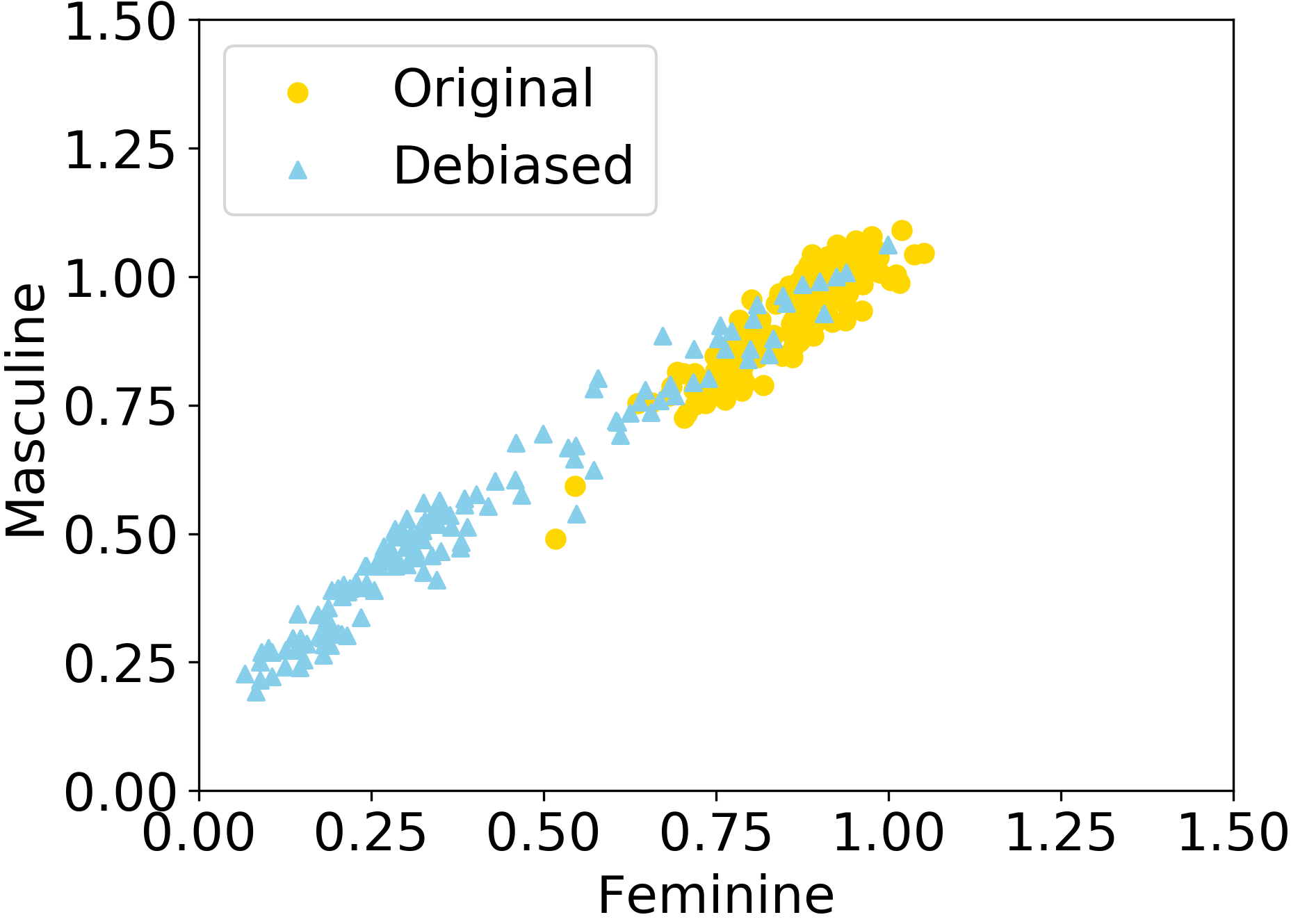}
		\caption{ELECTRA}
	\end{subfigure}
\caption{Scatter plot of gender information of hidden states for original and debiased stereotype words.}
\label{fig:plots}
\end{figure*}

Following \newcite{dev2020on-measuring}, we use the multi-genre co-reference-based natural language inference (MNLI) dataset for evaluating gender bias.
This dataset contains sentence triples where a premise must be neutral in entailment w.r.t. two hypotheses.  
If the predictions made by a classifier that uses word embeddings as features deviate from neutrality, it is considered as biased.
Given a set containing $M$ test instances, let the entailment predictor's probabilities for the $m$-th instance for entail, neutral and contradiction labels be respectively $e_{m}$, $n_{m}$ and $c_{m}$.
Then, they proposed the following measures to quantify the bias:
(1) Net Neutral (\textbf{NN}): ${\rm NN} = \frac{1}{M} \sum_{m=1}^{M} n_{m}$;
(2) Fraction Neutral (\textbf{FN}): ${\rm FN} = \frac{1}{M} \sum_{m=1}^{M} \textbf{1} [{\rm neutral} = {\rm max}(e_{m}, n_{m}, c_{m})]$; and
 (3) Threshold $\tau$ (\textbf{T:$\tau$}): T:$\tau$ $= \textbf{1} [n_{m} \geq \tau]$, where we used $\tau = 0.7$ following  \newcite{dev2020on-measuring}.
For an ideal (bias-free) embedding, all three measure would be 1.

In \autoref{tbl:snli_size}, we compare our proposed method against the \emph{noncontextualised debiasing} method proposed by \newcite{dev2020on-measuring} where they debias Layer 1 of BERT-large model using an orthogonal projection to the gender direction during training and evaluation.
In addition to the above-mentioned measures, we also report the entailment accuracy on the matched (in-domain) and mismatched (cross-domain) denoted respectively by \textbf{MNLI-m} and \textbf{MNLI-mm} in  \autoref{tbl:snli_size} to evaluate the semantic information preserved in the embeddings after debiasing.

We see that the proposed method outperforms  noncontextualised debiasing~\cite{dev2020on-measuring}  in NN and T:0.7, and its performance of the MNLI task is comparable to the original embeddings.
This result further confirms that the proposed method can not only debias well but can also preserve the pre-trained information.
Moreover, it is consistent with the results reported in \autoref{tbl:seat_gleu} and shows that debiasing all layers is more effective than only the first layer as done by \newcite{dev2020on-measuring}.

\subsection{The Importance of Debiasing All Layers}

In \autoref{tbl:seat_gleu}, we investigated the bias for the final layer, but it is known that the contextualised embeddings are learned at each layer~\cite{bommasani-etal-2020-interpreting}.
Therefore, to investigate whether by debiasing in each layer we are able to remove the biases of the entire contextualised embeddings, 
we evaluate the debiased embeddings at each layer on SEAT  6, 7, 8 datasets and report the averaged metrics for \textbf{all-token}, \textbf{first-token}
and \textbf{last-token} methods in \autoref{tbl:layer_seat}.
We see that, on average, \textbf{fitst-token} and \textbf{last-token} methods have more bias than \textbf{all-token}.
Therefore, we conclude that It is not enough to debias only the first and last layers even in DistilBERT, which has a small number of layers.
These results show that biases in the entire contextualised embedding cannot be reliably removed by debiasing only some selected layers, but rather the importance of debiasing all layers consistently.



\subsection{Visualizing Debiasing Results}

To further illustrate the effect of debiasing using the proposed \textbf{all-token} method, we visualise the similarity scores of a stereotypical word with feminine and masculine dimensions as follows.
First, for each target word $t$, its hidden state, $E_{i}(t, x)$ in the $i$-th layer of the model $E$ in a sentence $x$ is computed.
Next, we average those hidden states across all sentences in the dataset that contain $t$ to obtain $\hat{E}_{i}(t) = \frac{1}{|\cT|}\sum_{x \in \cT} E_{i}(t,x)$.
Likewise, we compute $\hat{E}_{i}(f)$ and $\hat{E}_{i}(m)$ respectively for each feminine ($f$) and masculine ($m$) word.
Next, we compute, $s_{i}^{f}$, the cosine similarity between each $\hat{E}_{i}(f)$ and the feminine vector $\vec{v}_{i}(f)$, and the cosine similarity, $s_{i}^{m}$, between each $\hat{E}_{i}(f)$ and the masculine vector $\vec{v}_{i}(f)$.
$\vec{s}_{i}^{f}$ and $\vec{s}_{i}^{m}$, respectively, are averaged over all layers in a contextualised embedding model
to obtain $\vec{s}_{\rm Avg}^{f}$ and $\vec{s}_{\rm Avg}^{m}$, which represent how much gender information each gender word contains on average.

We then compute the cosine similarity, $\vec{s}_{i}^{t,f}$, between each stereotype word's averaged embedding, $\hat{E}_{i}(t)$ and the feminine vector $\vec{v}_{i}(f)$. 
Similarly, we compute the cosine similarity $\vec{s}_{i}^{t,m}$ between each stereotype word's averaged embedding $\hat{E}_{i}(t)$ and the masculine vector $\vec{v}_{i}(m)$.
We then average $\vec{s}^{t,f}$ and $\vec{s}^{t,m}$ over the layers in $E$ respectively, to compute $\vec{s}_{\rm Avg}^{t,f}$ and $\vec{s}_{\rm Avg}^{t,m}$, which represent how much gender information each stereotype word contains on average.
Finally, we visualise the normalised female and male gender scores given respectively by $\vec{s}_{\rm Avg}^{t,f} / \vec{s}_{\rm Avg}^{f}$ and $\vec{s}_{\rm Avg}^{t,m} / \vec{s}_{\rm Avg}^{m}$.
For example, a zero $\vec{s}_{\rm Avg}^{t,f} / \vec{s}_{\rm Avg}^{f}$ value indicates that $t$ does not contain female gender related information, whereas a value of one indicates that it contains all information about the female gender.
\autoref{fig:plots} shows each stereotype word with its normalised female ad male gender scores respectively in $x$ and $y$ axises.
For a word, a yellow circle denotes its original embeddings, and the blue triangle denotes the result of debiasing using the \textbf{all-token} method.

We see that with the original embeddings, stereotypical words of are distributed close to one, indicating that they are highly gender-specific.
On the other hand, we see that the debiased BERT, DistilBERT and ELECTRA have similar word distributions compared to the original embeddings respectively, with an overall movement towards zero.
On the other hand, for RoBERTa, debiased embeddings are mainly distributed from zero to around one compared to the original embeddings.
Moreover, for ALBERT, the debiased embeddings are close to zero, but unlike the original distribution, the debiased embeddings are mainly clustered around zero.
This shows that RoBERTa and ALBERT do not retain structure of the original distribution after debiasing. 
While ALBERT over-debiases pre-trained embeddings of stereotypical words, RoBERTa under-debiases them.
This trend was already confirmed on the downstream evaluation tasks conducted in \autoref{tbl:seat_gleu}.

\section{Conclusion}

We proposed a debiasing method for pre-trained contextualised word embeddings, operating at token- or sentence-levels. 
Our experimental results showed that the proposed method effectively debiases discriminative gender-related biases,
 while preserving useful semantic information in the pre-trained embeddings.
The results showed that the downstream task was more effective in debias than the previous studies.

\bibliography{contdebias}
\bibliographystyle{acl_natbib}

\end{document}